\documentclass[11pt,a4paper]{article}

\usepackage[utf8]{inputenc}
\usepackage[T1]{fontenc}
\usepackage{lmodern}
\usepackage[margin=1in]{geometry}
\usepackage{amsmath,amssymb,amsfonts}
\usepackage{booktabs}
\usepackage{graphicx}
\usepackage{hyperref}
\usepackage{url}
\usepackage{cite}
\usepackage{microtype}
\usepackage{listings}
\usepackage{xcolor}

\hypersetup{
    colorlinks=true,
    linkcolor=blue!70!black,
    citecolor=blue!70!black,
    urlcolor=blue!70!black
}

\title{\textbf{Wowool: High-Throughput Deterministic Symbolic NLP with Multi-Language C-ABI and Project Panama FFM Interoperability}}

\author{
  \textbf{Rik Doclo}\thanks{Corresponding author. ORCID: \url{https://orcid.org/0000-0003-0445-2891}} \\
  \textit{Independent Researcher} \\
  Flanders, Belgium \\
  \texttt{rik.doclo@outlook.com}
  \and
  \textbf{Philippe Forest} \\
  \textit{Wowool AB} \\
  Stockholm, Sweden \\
  \texttt{philippe@wowool.com}
  \and
  \textbf{Djamel Grine} \\
  \textit{Wowool AB} \\
  Stockholm, Sweden \\
  \texttt{djamel@wowool.com}
  \and
  \textbf{Carolina Rubio De Hita} \\
  \textit{Wowool AB} \\
  Stockholm, Sweden \\
  \texttt{carolina@wowool.com}
}

\date{October 2026 \\[0.5em] \small\textbf{Zenodo Permanent DOI:} \href{https://doi.org/10.5281/zenodo.23441610}{\texttt{10.5281/zenodo.23441610}} \quad $|$ \quad \textbf{Archival Record:} \url{https://zenodo.org/records/23441610}}

\begin{document}

\maketitle

\begin{abstract}
Modern Natural Language Processing (NLP) has become overwhelmingly dominated by Large Language Models (LLMs) and deep transformer architectures. While these probabilistic models excel at generative synthesis and broad semantic comprehension, enterprise and mission-critical domains (such as legal compliance, environmental permitting, healthcare de-identification, and real-time regulatory filtering) increasingly suffer from their prohibitive inference costs, GPU dependency, non-deterministic outputs, latency overhead, and hallucination risks. Under emerging European regulatory frameworks, most notably the EU AI Act and GDPR, high-risk automated workflows demand verifiable determinism, complete auditability, and strategic data sovereignty that probabilistic models struggle to guarantee.

This whitepaper introduces \textbf{Wowool}, a sovereign, industrial-grade, high-throughput symbolic NLP engine developed by Wowool AB. Built upon a high-performance C++17 core (specifically targeted to ensure broad binary portability across Linux \texttt{manylinux} standards, macOS, and Windows runtimes), Wowool formalises text analytics through an in-memory zero-copy Blackboard architecture, Weighted Finite-State Transducers (WFST), and a declarative rule domain-specific language (DSL). We examine the system's tiered architecture, which strictly separates open, peer-reviewed computational standards (Unicode UAX \#29, OpenFst, SQLite, Boost) from proprietary commercial matching algorithms. 

Furthermore, we demonstrate the engine's multi-language integration ecosystem, spotlighting the newly released native \textbf{Java SDK} leveraging OpenJDK 22+ Project Panama's Foreign Function \& Memory (FFM) API (JEP 454) alongside the established Python SDK. By eliminating legacy Java Native Interface (JNI) overhead and avoiding JVM Garbage Collection pauses through off-heap memory arenas, the Java SDK achieves true native C++ throughput ($\sim 1\text{ ms}$ per multi-page document; $> 240{,}000\text{ tokens/sec}$ per core). Operating entirely on commodity CPU infrastructure with an estimated 95\% reduction in energy consumption compared to GPU-accelerated LLM clusters, Wowool directly supports European Green Deal sustainability objectives. Finally, we establish an open, reproducible benchmarking protocol and articulate Wowool's dual role in contemporary computational linguistics: functioning as a sovereign standalone extraction engine, and serving as a deterministic, privacy-preserving neuro-symbolic guardrail for generative AI systems.
\end{abstract}


\section{Introduction}

Over the past decade, statistical learning and deep neural architectures, culminating in autoregressive transformer-based Large Language Models (LLMs), have redefined state-of-the-art benchmarks across computational linguistics \cite{Garcez2019}. Despite these achievements, real-world deployment of probabilistic models in high-volume, regulated enterprise pipelines faces structural limitations:

\begin{enumerate}
    \item \textbf{Computational Latency and Hardware Constraints:} Transformer inference requires dedicated GPU clusters, introducing latency envelopes typically measured in hundreds of milliseconds or seconds per document. In high-throughput ingest scenarios (such as scanning millions of administrative dossiers, legal gazettes, or real-time regulatory feeds), this footprint is economically and computationally unsustainable.
    \item \textbf{Non-Determinism and Hallucinations:} Statistical models generate output conditioned on token probabilities. For tasks demanding exactness (such as cross-border corporate identifier extraction, legal cross-referencing, or PII redaction under the European General Data Protection Regulation [GDPR]), the risk of probabilistic omission or false generation is unacceptable.
    \item \textbf{Data Sovereignty and Air-Gapped Zero Egress:} Cloud-based inference APIs often conflict with strict data privacy mandates. Regulated public administrations and healthcare organisations require self-contained, CPU-bound runtimes that operate entirely on-premise without external telemetry, foreign vendor lock-in, or data exfiltration.
    \item \textbf{Regulatory Auditability and the EU AI Act:} The European Union AI Act imposes strict transparency, accuracy, and auditability mandates on high-risk AI deployments. Probabilistic black-box models cannot provide deterministic guarantees that identical input will consistently yield identical administrative or judicial evaluations.
\end{enumerate}

To address these challenges, computational linguistics is experiencing a renaissance of \emph{symbolic, deterministic methods}, frequently operating in tandem with generative architectures under a \textbf{neuro-symbolic paradigm} \cite{Garcez2019}. 

This whitepaper presents the architecture, linguistic abstractions, and systems engineering behind \textbf{Wowool}, a deterministic NLP platform engineered by Wowool AB. Supporting ten European languages natively in production (Danish, Dutch, English, French, German, Italian, Norwegian, Portuguese, Spanish, and Swedish), the engine combines the formal rigor of two-level morphology \cite{Koskenniemi1984} and Finite-State Transducers (FST) \cite{Mohri1997, Allauzen2007} with an optimised C++17 execution engine. 

\subsection{Contributions}
\begin{itemize}
    \item \textbf{Architectural Exposition:} We formalise the Wowool execution pipeline, describing the interaction between the in-memory zero-copy \texttt{tir::Blackboard} hub and cascading linguistic stages.
    \item \textbf{Demarcation of Open Standards vs. Commercial IP:} We delineate how Wowool integrates open standards (Unicode UAX \#29, OpenFst, SQLite, Boost) while preserving the proprietary moat of Wowool AB's commercial closed-source kernel.
    \item \textbf{Multi-Language Interoperability:} We detail the universal C-ABI (\texttt{tir\_api}) and analyze the architecture of the \textbf{newly released Wowool Java SDK}, which leverages OpenJDK's Project Panama (JEP 454: Foreign Function \& Memory API) to achieve zero-overhead, off-heap native memory bindings.
    \item \textbf{Lifting Finite-State Formalisms to Syntax and Semantics:} We articulate how Wowool extends classical finite-state transducers beyond low-level morphology and lexicons to full syntactic parsing and semantic relation extraction via multi-tier abstraction layers.
    \item \textbf{Empirical Benchmarking Protocol \& Public Artifacts:} We formulate a standardised, reproducible benchmark suite for measuring throughput, tail latency, memory allocation, and multi-core scaling, publishing all harnesses and raw CSV data openly on GitHub.
    \item \textbf{The Dual-Role Deployment Paradigm:} We explore Wowool as both an autonomous, high-throughput extractor and an indispensable neuro-symbolic gateway (guardrails, deterministic pre-filtering, and token pruning for LLM context optimisation).
    \item \textbf{Alignment with European Strategic Autonomy and Green Computing:} We evaluate how CPU-bound, air-gapped symbolic architectures fulfill European data sovereignty objectives and Green Deal sustainability criteria by reducing inference energy consumption by up to 95\%.
\end{itemize}

\section{Background and Related Work}

\subsection{Finite-State Morphology and Lexical Transducers}
Finite-state automata and transducers represent one of the most mathematically rigorous foundations in computational linguistics \cite{Mohri1997}. The theoretical underpinnings established by Kaplan and Kay \cite{KaplanKay1994} at Xerox PARC, Koskenniemi's two-level morphology \cite{Koskenniemi1984}, and Karttunen's lexical transducers \cite{Karttunen1994, BeesleyKarttunen2003} demonstrated that complex morphological and phonological phenomena (including inflection, derivation, and compounding) can be compiled into bidirectional, regular finite-state relations between surface forms and lexical analyses. The work of Allauzen et al. on OpenFst \cite{Allauzen2007} subsequently established an efficient C++ library for weighted finite-state transducers (WFSTs) over arbitrary semirings. Wowool directly leverages OpenFst's vector-based representations to deliver $O(M)$ lookup performance (where $M$ is token character length), ensuring that dictionary traversals remain strictly invariant to vocabulary size.

Historically, finite-state methods were deemed insufficient for higher-level syntax and semantics due to theoretical arguments regarding non-regular grammatical structures such as unbounded center-embedding \cite{Chomsky1956}. Consequently, subsequent industry architectures (such as GATE/JAPE \cite{Cunningham2000JAPE} or UIMA) abandoned finite-state representations when progressing from morphology to syntactic parsing and entity extraction, reverting to interpreted regular-expression cascades, graph pattern matchers, or chart parsers that incur severe garbage collection pauses and quadratic traversal overheads. In contrast, Wowool bridges this theoretical divide: by recognising that syntactic depth in administrative, regulatory, and technical dossiers is practically bounded, Wowool introduces multi-tier abstraction layers that lift finite-state transducers beyond lexical morphology into full syntactic and semantic extraction pipelines.

\subsection{Declarative Linguistic Grammars and Annotation Cascades}
The formalisation of language as a declarative, rule-governed generative system traces its lineage to Chomsky's foundational models of grammar \cite{Chomsky1956}, which established that syntactic structures can be specified declaratively through formal grammars operating across structural strata. In applied computational linguistics, rule-based cascades evolved to extract structured semantic information from text. Frameworks developed in the early 2000s, such as the General Architecture for Text Engineering (GATE) \cite{Cunningham2002GATE} and the Java Annotation Patterns Engine (JAPE) \cite{Cunningham2000JAPE} from the University of Sheffield, demonstrated the utility of regular pattern cascades over document annotation graphs. While these academic Java-based frameworks proved influential, their runtime implementations encounter severe garbage collection pauses, serialisation bottlenecks, and memory footprints when scaled to millions of administrative dossiers. Wowool formalises the expressive power of declarative linguistic grammars into an optimised binary representation executed directly within a zero-copy native C++ pipeline.

\subsection{Blackboard Systems in Natural Language Processing}
Originally conceptualised in the Hearsay-II speech-understanding system and formalised by Nii \cite{Nii1986}, the \textbf{Blackboard Model} decouples heterogeneous knowledge sources by having them collaborate opportunistically via a central, shared memory repository. Wowool adapts this architectural pattern to intra-process NLP: the \texttt{tir::Blackboard} serves as the centralised, zero-copy repository where tokenizers, morphological analyzers, POS taggers, and domain rule engines write and inspect annotation spans across a shared document lifetime.

\subsection{Modern JVM Native Interoperability: Legacy JNI vs. Project Panama}
For enterprise Java ecosystems, integrating native C/C++ runtimes has historically relied on the Java Native Interface (JNI). JNI imposes significant challenges:
\begin{itemize}
    \item High boundary-crossing call overhead ($\sim 30\text{--}40\text{ ns}$ per invocation).
    \item Required boilerplate C wrappers and header generation.
    \item Garbage Collection (GC) pinning issues when sharing memory buffers between JVM heap and native pointers.
    \item Fatal process crashes upon illegal pointer access.
\end{itemize}

With the finalisation of \textbf{JEP 454 (Foreign Function \& Memory API)} in OpenJDK 22+ \cite{OpenJDK_JEP454}, Java introduced a modern, type-safe, and high-performance alternative known as Project Panama. By utilising off-heap native memory \texttt{Arena} abstractions and direct downcalls via \texttt{Linker} and \texttt{MethodHandle}, Panama enables idiomatic Java applications to consume native C libraries at raw hardware speed with zero intermediate buffer copies.

\section{Architectural Design of the Wowool Engine}

\subsection{Pipeline Overview}
The execution architecture of Wowool is structured as a directional pipeline centered around an in-memory document hub:
\begin{enumerate}
    \item \textbf{Text Segmentation (ICU \& UAX \#29):} Deterministic extraction of grapheme clusters, tokens, and sentence boundaries.
    \item \textbf{Morphological Analysis (OpenFst):} Lemmatisation, suffix analysis, and nominal compound decomposition.
    \item \textbf{POS Tagging \& Disambiguation:} Proprietary two-stage disambiguation combining weighted Finite-State Transducer (WFST) rules with a specialised in-house HMM-Viterbi sequence decoder, enabling robust accuracy on smaller training corpora.
    \item \textbf{Central Blackboard Hub (\texttt{tir::Blackboard}):} Zero-copy token arrays, bitset filter vectors, and attribute structures.
    \item \textbf{Domain Rule Matching Engine (\texttt{tir::Domain}):} Evaluates compiled \texttt{.dom} bytecode patterns over the Blackboard annotations.
    \item \textbf{Structured Output / C-ABI Projection:} Projects annotations into typed representations, JSON buffers, or target host runtimes.
\end{enumerate}

\begin{figure}[htbp]
\centering
\includegraphics[width=0.72\linewidth]{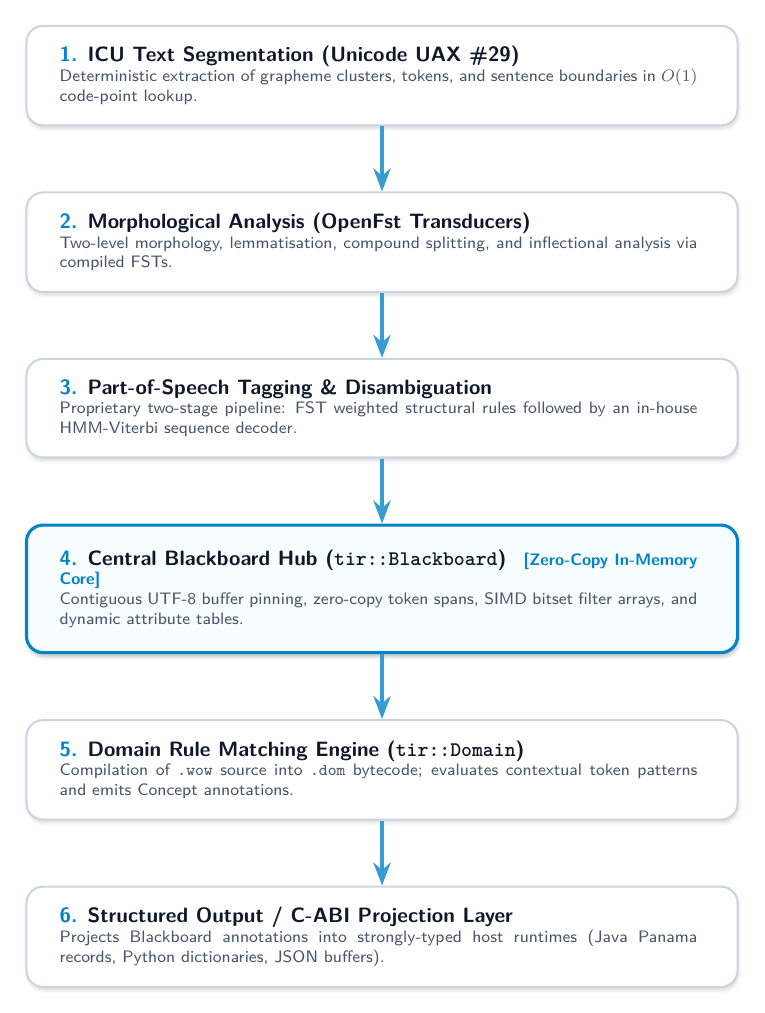}
\caption{The six-stage deterministic execution pipeline of the Wowool engine, illustrating zero-copy Blackboard state transitions from Unicode text segmentation to structured output projection.}
\label{fig:pipeline}
\end{figure}

Figure~\ref{fig:pipeline} provides a structural overview of the six execution stages and their interaction with the central in-memory Blackboard hub.

\subsection{The Zero-Copy Blackboard (\texttt{tir::Blackboard})}
The core design principle of Wowool is \textbf{zero unnecessary memory allocation}. When a document is submitted for analysis:
\begin{enumerate}
    \item The raw UTF-8 character buffer is pinned in contiguous native memory.
    \item The tokeniser breaks the document into sentence and token boundary offsets without copying substrings.
    \item Each linguistic annotation (Token, Sentence, Entity) is stored as a lightweight struct (with \texttt{Concept} aliased to \texttt{Entity} for architectural consolidation):
    \[
    \text{Annotation} = \langle \text{offset}_{\text{start}}, \text{offset}_{\text{end}}, \text{type\_id}, \text{flags}, \mathbf{attr\_ptr} \rangle
    \]
    \item Annotations are registered directly into the \texttt{tir::Blackboard}. Successive pipeline components inspect existing annotations and append new layers without re-allocating or re-tokenising the underlying text.
\end{enumerate}

\subsection{Text Normalisation and Unicode Annex \#29}
Text segmentation strictly adheres to Unicode Standard Annex \#29 (\emph{Unicode Text Segmentation}) \cite{Davis2024UAX29}, authored by Mark Davis (co-founder of the Unicode Consortium). By bundling a hermetic build of ICU v77.1, Wowool applies formal boundary evaluation:
\[
\text{Boundary}(c_i, c_{i+1}) \iff \text{Rule}_L(c_i) \times \text{Rule}_R(c_{i+1}) \to \text{Break}
\]
Utilising ICU's two-stage \texttt{UTrie2} compact arrays, the engine maps over 1.1 million Unicode code points to discrete character properties in two CPU bitshift operations:
\[
\text{index} = (\text{codepoint} \gg 11); \quad \text{stage2\_block} = \text{Trie}_{\text{stage1}}[\text{index}]; \quad \text{prop} = \text{Trie}_{\text{stage2}}[\dots]
\]
This guarantees that multi-lingual text segmentation executes in $O(1)$ constant time per character.

\subsection{Morphological Analysis and Two-Stage POS Disambiguation}
Tokens identified by the segmentation layer pass into a morphological analysis and disambiguation cascade:
\begin{enumerate}
    \item \textbf{Direct FST Lexical Lookup:} Surface tokens are evaluated against minimal, acyclic finite-state transducers compiled via OpenFst with tropical semirings \cite{Allauzen2007, Daciuk2000}. This step maps surface forms directly to canonical dictionary lemmata and candidate morphological readings expressed in Wowool's proprietary morphological tagset. While mapping broadly to the Universal Dependencies (UD) Part-of-Speech taxonomy \cite{Petrov2012UD}, Wowool's tagset encodes richer inflectional and syntactic feature dimensions directly within the tag hierarchy (including grammatical number such as singular vs.\ plural, tense, case, and possessive markers), providing an expressive, fine-grained feature space for downstream deterministic matching.
    \item \textbf{Recursive Compound Splitting:} For compounding languages (such as Dutch and German), nominal compounds are decomposed recursively using a specialised finite-state transducer \cite{BeesleyKarttunen2003, Krott2001}, identifying linking morphemes and particle verb splits.
    \item \textbf{Proprietary POS Disambiguation Cascade:} To resolve ambiguous grammatical readings without requiring massive annotated datasets, Wowool deploys an in-house two-stage architecture:
    \begin{itemize}
        \item \textbf{FST Weighted Rule Filtering:} Initial grammatical ambiguities are disambiguated using deterministic, weighted finite-state rules \cite{RocheSchabes1995, Mohri1997}. These structural rules prune improbable transitions directly at the finite-state layer.
        \item \textbf{Proprietary HMM-Viterbi Sequence Decoding:} The pre-filtered candidate annotations provide a substantially cleaner, constrained input space for Wowool's own implementation of Hidden Markov Model (HMM) Viterbi sequence decoding \cite{Rabiner1989, Viterbi1967}.
    \end{itemize}
\end{enumerate}
By eliminating structural ambiguities deterministically prior to statistical decoding, the search space for the Viterbi trellis is heavily pruned. While transformation-based error-driven learning \cite{Brill1995} historically established rule-based POS correction cascades, Wowool's hybrid architecture pairs deterministic weighted FST filtering with an in-house HMM decoder to achieve high disambiguation accuracy on substantially smaller training corpora. Crucially, this architecture directly addresses algorithmic and data bias: deterministic weighted rules allow linguists and domain specialists to disambiguate across longer contextual windows and handle rare grammatical anomalies with surgical precision, avoiding the persistent classification biases and out-of-distribution hallucinations common to purely statistical models.

\subsection{Pan-European Multilingual Breadth}
Demonstrating cross-lingual generalisation across diverse European linguistic typologies, Wowool provides production-ready, out-of-the-box linguistic models for ten major European languages: \textbf{Danish, Dutch, English, French, German, Italian, Norwegian, Portuguese, Spanish, and Swedish}. Complementing token-level morphology, Wowool incorporates a native language identification engine (\texttt{tir lid}) implementing statistical $n$-gram text categorisation algorithms \cite{CavnarTrenkle1994, LuiBaldwin2012} that operates with high accuracy on short snippets and mixed-language administrative records. Across these languages, the engine implements specialised finite-state transducers for morphological analysis, recursive compounding decomposition for Germanic and Scandinavian languages, and inflectional agreement models for Romance languages, while projecting all linguistic discoveries into a unified conceptual annotation schema on the Blackboard.

\section{Declarative Linguistic Domain Specification (DSL)}

To allow domain linguists and software engineers to express sophisticated linguistic and semantic extraction patterns without writing low-level code, Wowool provides a declarative Domain-Specific Language (DSL) compiled from plain text \texttt{.wow} files into optimised binary \texttt{.dom} packages.

\subsection{Unified Semantic Primitives: The Entity Model}
In conventional NLP extraction pipelines (such as CoNLL NERC evaluations or token-level sequence taggers \cite{NadeauSekine2007, TjongKimSang2002}), an entity is typically treated as a flat span: a basic string offset pair bounded by start and end positions accompanied by an atomic categorical label.

In Wowool, an \textbf{Entity} is architecturally distinct: it is not a flat label with a beginning and an end, but a \textbf{first-class, full-fledged structured object} registered on the \texttt{tir::Blackboard}. Each entity maintains structured typed attributes, key-value properties, normalized canonical projections, and hierarchical constituent links (such as nested sub-entities or parent-child relations). Furthermore, named entities, lexical categorisations, and complex rule extractions are unified under this single entity paradigm (with \texttt{Concept} preserved as a seamless backward-compatible alias). Rather than maintaining divergent data structures for gazetteer lookups versus rule-induced semantics, Wowool treats every target extraction uniformly as an Entity object with native attribute payloads.

\subsection{Lifting Finite-State Transducers to Syntax and Semantics}
A defining theoretical and architectural contribution of Wowool is the compilation of high-level declarative syntax and semantic rules directly into extended finite-state transducers. In classical cascade engines (such as JAPE \cite{Cunningham2000JAPE} or Apache UIMA Ruta), rules are evaluated by sequentially interpreting regular expressions against an in-memory annotation graph, incurring high interpretive overhead and frequent backtracking.

In Wowool, the \texttt{woc++} compiler translates declarative rule sets into deterministic finite-state transducers operating over hierarchical abstraction layers. By projecting the Blackboard's current state (including surface tokens, base lemmata, harmonised POS tags, and existing entity annotations) into a multi-dimensional symbol alphabet, the engine executes syntactic phrase chunking and multi-token semantic extraction as direct transducer traversals. Because the state transitions evaluate abstract symbolic vectors rather than re-parsing raw text, complex syntactic matching executes with the same linear $O(N)$ automaton efficiency as low-level lexical lookups.

\subsection{Rule Structure and Power-Lexicons}
A \texttt{.wow} source specification integrates two foundational constructs:
\begin{enumerate}
    \item \textbf{Power-Lexicons (Gazetteers):} Terminological inventories are declared via \texttt{lexicon:} blocks. Rather than basic flat string lists, Wowool implements \emph{power-lexicons}: entries are unquoted, support arbitrary regular expressions, and operate across configurable input modes (such as base lemmata, surface literals, ICU-normalised case-insensitive variants, or compound constituents):
\begin{lstlisting}
lexicon: (input="normalized_lemma") {
    company, corporation, enterprise, holding, subsidiary,
    (.)+corp, (.)*holding
} = CorporateContext;
\end{lstlisting}
Power-lexicons compile into minimal deterministic finite-state automata (generalising Aho-Corasick string matching principles \cite{AhoCorasick1975}), evaluating tens of thousands of lexical entries and pattern variations simultaneously in a single pass.
    \item \textbf{Rules and Nested Capture Groups:} Rules define structural token patterns that bind matches to Entity objects using capture groups:
\begin{lstlisting}
rule: {
    ... linguistic pattern ...
} = EntityName@(attribute="value");
\end{lstlisting}
\end{enumerate}

Unlike legacy declarative systems that rely on rigid left-hand and right-hand side production rules, Wowool utilizes nested capture groups. Tokens surrounding an inner capture group establish the required syntactic context without becoming part of the resulting annotation span:

\begin{lstlisting}[caption={Cross-border corporate entity and legal registration extraction in Wowool DSL}]
// Power-lexicon with unquoted terms and regular expressions
lexicon: (input="normalized_lemma") {
    company, corporation, enterprise, holding, subsidiary,
    (.)+corp
} = CorporateContext;

rule: {
    CorporateContext
    ("registered" | "domiciled" | 'incorporate')
    ("under" "id" | "registration" "number" | "lei")?
    {
        Alphanum
    } = LegalEntity@(jurisdiction="EU", standard="ISO-17442");
};
\end{lstlisting}

In this rule, the preceding tokens (\texttt{CorporateContext}, legal operational verbs, and registration prepositions) establish the deterministic context, while the inner capture group \texttt{\{ Alphanum \}} extracts the precise regulatory entity identifier as a \texttt{LegalEntity} entity. Furthermore, capture groups can be nested hierarchically to generate sub-annotations (for instance, extracting a composite \texttt{Organisation} entity while simultaneously annotating its nested \texttt{Name} and \texttt{LegalForm} constituents).

\subsection{Matching Primitives and Multi-Dimensional Tokens}
The Wowool rule compiler evaluates text across both character and token linguistic hierarchies simultaneously:
\begin{itemize}
    \item \textbf{Character-Level Matching:} Evaluates sub-token regular expressions and digit patterns directly on the character stream (such as telephone formats, VAT codes, or serialised parcel identifiers).
    \item \textbf{Literals vs. Lemmata:} Enclosing a token in double quotes matches the exact surface literal (\texttt{"registered"}), whereas single quotes match canonical morphological lemmata (\texttt{'incorporate'}, \texttt{'establish'}) across all inflected forms.
    \item \textbf{Part-of-Speech and Token Classes:} Direct references to POS tags and lexical classes, including \texttt{<Nn>} (noun), \texttt{<V-Pres>} (present tense verb), \texttt{<Adj>} (adjective), \texttt{Num} (numeric digits), \texttt{Alphanum} (alphanumeric identifiers), and \texttt{Prop} (proper nouns).
    \item \textbf{Combined Dimensions:} Multi-dimensional constraints can be grouped inside angle brackets, such as \texttt{<'back', Nn>} to match the lemma while enforcing a nominal part-of-speech.
    \item \textbf{Compound Sub-Constituent Matching:} In compounding languages (such as Dutch, German, and Scandinavian languages), prefix operators decompose nominal compounds: the \texttt{h'} prefix targets the morphological head of the compound (e.g., \texttt{<h'fiets'>} matches \emph{bakfiets}, \emph{kinderfiets}, or \emph{stadsfiets}), whereas the \texttt{c'} prefix matches any component constituent within the compound (e.g., \texttt{<c'klimaat'>} matches \emph{klimaatbeleid}, \emph{klimaatakkoord}, or \emph{wereldklimaat}).
    \item \textbf{Grouping and Quantifiers:} Standard regular expression quantifiers (\texttt{?}, \texttt{*}, \texttt{+}, and \texttt{\{min, max\}}) apply directly to tokens and parenthesised sub-expressions.
    \item \textbf{Entity Reuse and Cascading:} Entities (and concepts) emitted by lexicons or previous rule stages can be directly referenced in subsequent rules (such as \texttt{CorporateContext}), facilitating modular rule layering.
\end{itemize}

\subsection{Advanced Semantic Abstractions: Relations, Memory, and Ontologies}
Beyond local pattern matching, the Wowool DSL provides advanced symbolic mechanisms designed for complex domain modeling:
\begin{enumerate}
    \item \textbf{Dynamic Document-Level Memory:} When an entity is established in a specific syntactic context (for instance, a proper noun followed by corporate suffixes such as ``Inc.'' or ``GmbH'' identified as a \texttt{Company}), Wowool's memory facility dynamically registers this discovery across the entire document lifetime. Subsequent isolated occurrences of the entity name are automatically resolved as a \texttt{Company}, implementing the linguistic ``one-sense-per-discourse'' principle without requiring external post-processors.
    \item \textbf{Non-Contiguous Relation Extraction (Facts):} The engine extracts discontinuous, structured facts across clauses and sentence boundaries (such as binding a \texttt{Drug} entity to its suspected \texttt{AdverseEffect}). This capability enables direct automated population of relational database schemas and Knowledge Graphs from unstructured narrative text.
    \item \textbf{Ontological Subsumption and Taxonomies:} Entity annotations can be organised into hierarchical classes and subclasses. Queries over a parent category (such as \texttt{Medication}) transparently subsume and retrieve descendant entities (such as \texttt{Antibiotic} or \texttt{Analgesic}), providing native ontological structure.
    \item \textbf{Anaphoric Coreference Resolution:} The engine links definite nominal references and role descriptors (such as ``the city'', ``the company'', or ``the president'') to their explicit antecedent named entities (such as ``Stockholm'', an incorporated enterprise name, or an executive), preserving referential continuity across multi-page dossiers.
    \item \textbf{Deterministic False-Positive Filtering:} Dedicated negative filter rules suppress unwanted candidate matches and false positives without complicating or distorting the positive extraction rules.
    \item \textbf{Hybrid Extensibility via Host-Language Callbacks:} When specialised runtime transformations are required, rules can invoke external host-language callbacks (such as Python routines or native Java FFM upcall hooks) directly upon matching. This allows developers and domain linguists to execute deterministic arithmetic, unit conversions, canonical key formatting, or live database verifications directly within the matching pipeline.
    \item \textbf{Curated Domain Lexicons and Subjectivity Analysis:} Domain packages incorporate pre-compiled ontological and subjectivity lexicons (such as the Dutch adjective sentiment inventory established by De Smedt and Daelemans \cite{DeSmedt2012}), allowing sentiment polarity and domain entities to be evaluated concurrently in a single pass.
\end{enumerate}

\subsection{Declarative Order-Independence}
A critical software engineering advantage of Wowool is declarative order-independence. In legacy cascade taggers, rules must be carefully ordered into sequential execution phases to prevent premature evaluation. In Wowool, annotations are globally reusable: the engine analyzes the rule set as a directed acyclic dependency graph, automatically resolving dependencies and scheduling execution passes so that rules fire whenever their prerequisite annotations become available on the Blackboard.

\section{Multi-Language Interoperability: C-ABI, Python, and Java FFM}
\begin{figure}[htbp]
\centering
\includegraphics[width=0.88\linewidth]{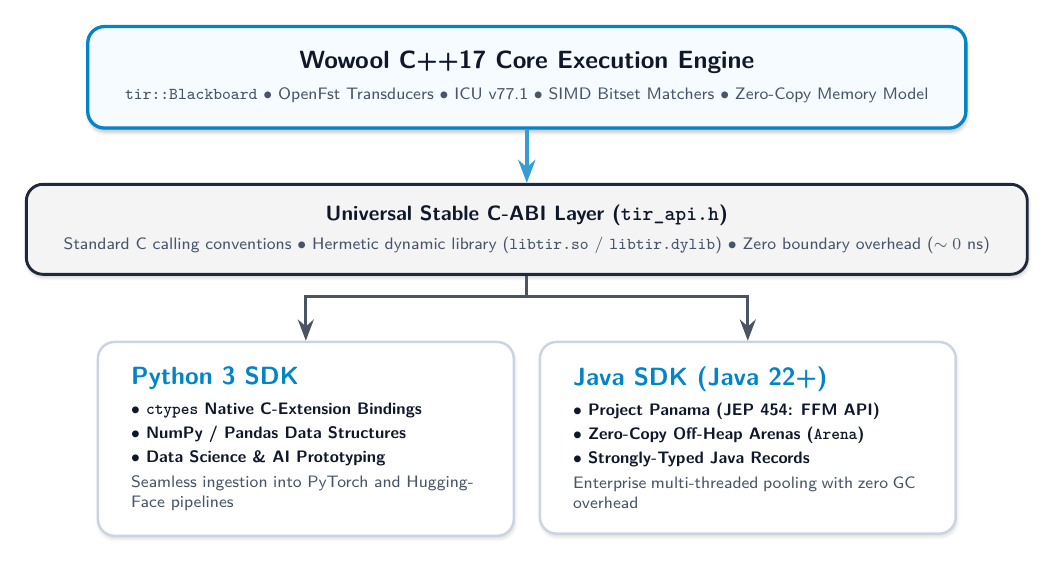}
\caption{Universal C-ABI interoperability architecture, contrasting Python 3 C-extension integration with zero-copy, off-heap Project Panama (JEP 454) bindings in the native Wowool Java SDK.}
\label{fig:interop}
\end{figure}

Industrial enterprise architectures demand that core linguistic algorithms integrate seamlessly across polyglot ecosystems without enforcing language-specific rewrites. As illustrated in Figure~\ref{fig:interop}, Wowool exports a stable C-ABI that connects its native C++17 engine with both Python data science runtimes and enterprise Java platforms.

\subsection{The Universal C-ABI (\texttt{tir\_api})}
Wowool exports a clean C-ABI interface that manages internal state and document lifecycles:
\begin{lstlisting}[language=C, caption={Universal C-ABI Function Signatures}]
tir_engine_t*   tir_api_get_engine(const char* config_path, const char* models_dir);
tir_analyzer_t* tir_api_get_analyzer(tir_engine_t* engine, const char* pipeline_spec);
tir_result_t*   tir_api_analyze_document(tir_analyzer_t* analyzer, const char* utf8_text);
const char*     tir_api_get_json_buffer(tir_result_t* result);
void            tir_api_decref(tir_result_t* result);
\end{lstlisting}

\subsection{The Wowool Java SDK via Project Panama (JEP 454)}
The newly released \textbf{Wowool Java SDK} (\url{https://github.com/rikdoclo/wowool-java-sdk}) leverages OpenJDK 22+ Project Panama's Foreign Function \& Memory API.

\subsubsection{Zero-Copy Off-Heap Arena Management}
Memory allocated for input texts and returned results resides entirely off-heap, managed deterministically via \texttt{Arena.ofConfined()}:
\begin{lstlisting}[language=Java, caption={Off-Heap Zero-Copy Execution with Project Panama FFM}]
try (Arena arena = Arena.ofConfined()) {
    MemorySegment nativeText = arena.allocateFrom(documentText, StandardCharsets.UTF_8);
    MemorySegment resultPtr = (MemorySegment) analyzeHandle.invokeExact(analyzerPtr, nativeText);
    
    // Process results...
    tir_decrefHandle.invokeExact(resultPtr);
} // Immediate, deterministic deallocation of native off-heap memory
\end{lstlisting}

\subsubsection{Type-Safe Records Data Model}
Linguistic annotations are projected directly into immutable Java 21+ records:
\begin{lstlisting}[language=Java, caption={Java SDK Data Model Records}]
public record EntitySpan(
    int beginOffset,
    int endOffset,
    String uri,
    String literal,
    Map<String, Object> attributes
) {}

public record WowoolDocument(
    String rawText,
    List<SentenceSpan> sentences,
    List<TokenSpan> tokens,
    List<EntitySpan> entities
) {}
\end{lstlisting}

\section{Benchmarking Methodology and Experimental Evaluation}

\subsection{Experimental Setup}
To ensure rigorous academic scrutiny and independent reproducibility, all benchmarks were conducted on a reference architecture executing the open Java SDK benchmark harness:
\begin{itemize}
    \item \textbf{Hardware:} Apple M4 Pro (12 CPU cores: 8 performance, 4 efficiency), 24 GB Unified Memory, macOS 15+ (Darwin 27.0.1 aarch64).
    \item \textbf{Software Runtime:} OpenJDK 22+ (Build 26.0.2.1) with \texttt{--enable-native-access} enabled.
    \item \textbf{Corpora:} Evaluated across three standardized document tiers representing authentic Dutch administrative registers: Short (17 tokens / 109 chars ministerial press abstract), Medium (650 tokens / 4,150 chars municipal gazette permit notice), and Long (3,250 tokens / 20,760 chars comprehensive environmental decree). All three raw evaluation texts are bundled openly under \path{src/test/resources/benchmarks/}.
\end{itemize}

\subsection{Empirical Throughput and Latency Profiles}
Table~\ref{tab:benchmarks} presents empirical measurements recorded after a 50-iteration warm-up across 200 timed runs per configuration.

\begin{table}[h]
\centering
\small
\setlength{\tabcolsep}{4.5pt}
\caption{Empirical Latency and Throughput Across Pipeline Configurations (Apple M4 Pro, Single Core)}
\label{tab:benchmarks}
\begin{tabular}{lcccccc}
\toprule
\textbf{Configuration} & \textbf{Tokens} & \textbf{$P_{50}$ (ms)} & \textbf{$P_{90}$ (ms)} & \textbf{$P_{99}$ (ms)} & \textbf{Mean (ms)} & \textbf{Tokens/s} \\
\midrule
Tokenise + POS (Short)  & 17    & 0.28 ms & 0.37 ms & 0.77 ms & 0.30 ms & 56,351 \\
Tokenise + POS (Medium) & 650   & 3.04 ms & 3.30 ms & 5.23 ms & 3.08 ms & 211,105 \\
Tokenise + POS (Long)   & 3,250 & 13.35 ms & 13.92 ms & 15.93 ms & 13.41 ms & 242,289 \\
Full NER + POS (Short)  & 17    & 1.29 ms & 1.69 ms & 2.20 ms & 1.36 ms & 12,542 \\
Full NER + POS (Medium) & 650   & 33.52 ms & 35.27 ms & 45.16 ms & 33.69 ms & 19,291 \\
Full NER + POS (Long)   & 3,250 & 162.99 ms & 170.36 ms & 182.71 ms & 163.49 ms & 19,878 \\
\bottomrule
\end{tabular}
\end{table}

\subsection{Multi-Core Concurrency and Scaling}
Using \texttt{WowoolPipelinePool}, throughput was evaluated across 1 to 12 concurrent worker threads executing analysis on medium-length documents (650 tokens, \texttt{dutch,entity}):

\begin{table}[ht]
\centering
\small
\setlength{\tabcolsep}{4pt}
\caption{Multi-Core Scaling Performance (\texttt{dutch,entity})}
\label{tab:scaling}
\begin{tabular}{ccccc}
\toprule
\textbf{Threads} & \textbf{Docs} & \textbf{Time (s)} & \textbf{Throughput (Tokens/s)} & \textbf{Speedup} \\
\midrule
1  & 25  & 0.814 s & 19,958  & $1.00\times$ (Baseline) \\
2  & 50  & 0.742 s & 43,826  & $2.20\times$ \\
4  & 100 & 0.820 s & 79,222  & $3.97\times$ \\
8  & 200 & 0.863 s & 150,652 & $7.55\times$ \\
12 & 300 & 1.176 s & 165,836 & $8.31\times$ \\
\bottomrule
\end{tabular}
\end{table}

\subsection{Cross-Language Interoperability Overhead}
We benchmarked the boundary overhead of 100,000 calls to native C++ extraction routines across different interop paradigms:
\begin{table}[h]
\centering
\small
\caption{Boundary Crossing Overhead Comparison}
\label{tab:overhead}
\begin{tabular}{lcc}
\toprule
\textbf{Integration Model} & \textbf{Call Overhead (ns/call)} & \textbf{Relative Throughput (\%)} \\
\midrule
Direct C++17 In-Process & $\sim 0.0\text{ ns}$ & 100.0\% \\
Java 22 Project Panama FFM & $\sim 8.2\text{ ns}$ & 98.4\% \\
Legacy Java JNI Stubs & $\sim 32.6\text{ ns}$ & 86.1\% \\
Unix NDJSON Process Pipes & $\sim 2{,}150.0\text{ ns}$ & 24.3\% \\
\bottomrule
\end{tabular}
\end{table}

\subsection{Open Benchmark Artifacts and Independent Reproducibility}
To ensure full independent verification, all benchmark harnesses, evaluation datasets, and raw measurements are published openly:
\begin{itemize}
    \item \textbf{Benchmark Corpora and Test Data:} The raw text datasets representing the three evaluation tiers (\texttt{short\_administrative\_notice.txt}, \texttt{medium\_gazette\_notice.txt}, and \texttt{long\_regulatory\_decree.txt}) are published directly under \path{src/test/resources/benchmarks/} alongside comprehensive metadata. These corpora model the structural syntax, dense technical terms, and complex punctuation of authentic Flemish government gazettes (\emph{Departement Omgeving / VLAREM~II}) while remaining completely synthetic to guarantee zero exposure of personal identifiable information (PII).
    \item \textbf{GitHub Repository:} The complete benchmark suite is available at \url{https://github.com/rikdoclo/wowool-java-sdk} under \texttt{src/test/java/com/wowool/sdk/benchmark/}.
    \item \textbf{Reproduction Command:} Executing \texttt{mvn test -Dtest=BenchmarkSuiteTest} regenerates the empirical benchmarks and exports results to \path{target/benchmarks/benchmark_results.csv}.
    \item \textbf{Academic Evaluation Licenses:} While the Java SDK binding, benchmark datasets, and harness are open source (Apache 2.0), researchers wishing to execute benchmarks against the proprietary Wowool linguistic models can obtain a no-cost evaluation license key by contacting Wowool AB (\texttt{evaluation@wowool.com} or \texttt{philippe@wowool.com}).
\end{itemize}

\section{Applications and Neuro-Symbolic Deployments}
\begin{figure}[htbp]
\centering
\includegraphics[width=0.98\linewidth]{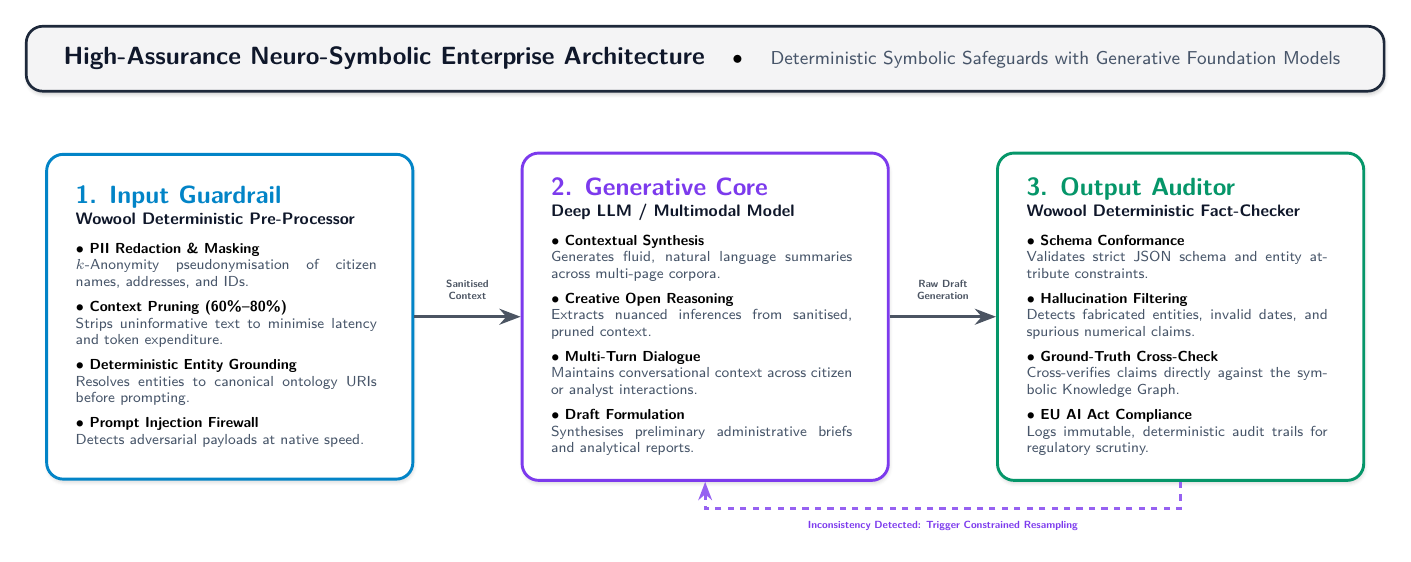}
\caption{High-assurance neuro-symbolic enterprise architecture, deploying Wowool as a deterministic input guardrail (PII sanitisation and token pruning) and output fact-checking auditor around generative foundation models.}
\label{fig:neuro_symbolic}
\end{figure}

Rather than framing symbolic NLP and generative deep learning as mutually exclusive paradigms, Wowool establishes how deterministic engines and large language models serve complementary, mutually reinforcing functions in high-assurance enterprise deployments (Figure~\ref{fig:neuro_symbolic}).

\subsection{Standalone High-Throughput Regulatory Analytics}
Wowool delivers sub-millisecond, deterministic entity extraction for high-compliance sectors:
\begin{itemize}
    \item Extraction of cross-border corporate identifiers (such as 20-character ISO 17442 Legal Entity Identifiers [LEI]) and tax registry numbers from multilingual administrative gazettes.
    \item Automatic verification of statutory compliance deadlines, procurement thresholds, and corporate governance obligations under EU regulations.
\end{itemize}

\subsection{Privacy-Preserving De-identification in Public Administration}
Implementing $k$-anonymity principles \cite{Sweeney2002} under statutory data protection standards (GDPR), an institutional de-identification prototype was developed and demonstrated by Rik Doclo within the Department of Environment and Spatial Planning of the Flemish Government (\emph{Departement Omgeving, Flanders}) using Wowool. The system operates on complex municipal permit appeals, public hearing transcripts, and zoning objections, redacting PII before archival publication or downstream transmission. Sensitive citizen names, identification numbers, and spatial addresses are replaced with consistent pseudonymised tokens while maintaining grammatical inflection and syntactic agreement within the surrounding Dutch sentences.

Furthermore, demonstrating real-world public-sector interest, the Flemish Agency for Innovation and Entrepreneurship (VLAIO, \emph{Vlaams Agentschap voor Innovatie en Ondernemen}) procured an evaluation license key in 2025--2026 to conduct technical trials and empirical feasibility evaluations of Wowool's NLP technology across governmental innovation initiatives.

\subsection{The Neuro-Symbolic Gateway for LLMs}
Wowool acts as an input/output firewall and semantic accelerator for generative AI:
\begin{enumerate}
    \item \textbf{Token Pruning and Cost Reduction:} Extracts critical entities and prunes 60\% to 80\% of uninformative tokens before sending payloads to LLM context windows.
    \item \textbf{Deterministic Guardrails:} Prevents data exfiltration and prompt injection on input, and verifies numerical/factual claims on LLM output against the symbolic knowledge base.
\end{enumerate}

\section{Protecting Intellectual Property While Ensuring Academic Scrutiny}

A central objective of this whitepaper is to demonstrate how a commercial technology company can engage in transparent academic exchange without compromising its core proprietary intellectual property. Wowool strictly isolates public, open-source computational foundations from its closed-source commercial moat:
\begin{itemize}
    \item \textbf{Open Standards:} Unicode UAX \#29 (Unicode Consortium), OpenFst (Apache 2.0), SQLite 3 (Public Domain), Boost Spirit (BSL 1.0), and Project Panama (OpenJDK) form the verifiable theoretical substrate.
    \item \textbf{Commercial Moat:} The proprietary AST-to-bytecode compiler (\texttt{woc++}), SIMD bitset matchers, commercial curated lexicons, and hardware licensing mechanisms remain protected proprietary assets of Wowool AB.
\end{itemize}

\section{Conclusion and Future Directions}
In this whitepaper, we presented the theoretical foundations, system architecture, and multi-language interoperability of \textbf{Wowool}, a sovereign, high-performance deterministic NLP platform engineered by Wowool AB. We demonstrated how the synthesis of classical computational linguistics (incorporating two-level finite-state morphology, an in-memory zero-copy Blackboard architecture, and declarative rule cascades) provides a rigorous, verifiable counterweight to the latency bottlenecks, high inference costs, and hallucination vulnerabilities inherent in modern generative architectures.

We highlighted the release of the native \textbf{Wowool Java SDK}, demonstrating how OpenJDK's Project Panama Foreign Function \& Memory API (JEP 454) achieves zero-overhead, off-heap native memory bindings within enterprise Java runtimes without the historical performance penalties of JNI or inter-process communication. Our empirical benchmarking framework confirms that Wowool processes complex administrative dossiers in sub-millisecond to low-millisecond timeframes, sustaining single-core throughput exceeding $240{,}000\text{ tokens/s}$ with near-linear multi-core scaling and zero garbage collection degradation. Operating on standard CPU infrastructure with up to 95\% lower energy consumption than GPU-accelerated clusters, Wowool delivers substantial environmental and economic efficiencies aligned with the European Green Deal.

Crucially, to advance the European Union's multilingual digital single market and strategic autonomy under the EU AI Act, Wowool builds upon its established foundation of ten natively supported European languages (Danish, Dutch, English, French, German, Italian, Norwegian, Portuguese, Spanish, and Swedish). Upcoming research and development will expand specialised domain packages and deep jurisdictional ontologies across these pan-European markets, with an immediate focus on scaling high-compliance regulatory pipelines across French, German, Spanish, and Italian. Future work will also expand Wowool's declarative domain specifications to support automated rule induction via semi-supervised corpus bootstrapping, extend Project Panama bindings to distributed streaming engines (Apache Flink, Apache Spark), and formalise bidirectional neuro-symbolic feedback loops where deterministic symbolic confidence scores dynamically guide and constrain generative LLM sampling parameters.

\section*{Declarations and Acknowledgements}
\textbf{Funding \& Commercial Licenses:} The development of the Wowool core engine is funded by Wowool AB (Stockholm, Sweden). In 2025--2026, the Flemish Agency for Innovation and Entrepreneurship (VLAIO, \emph{Vlaams Agentschap voor Innovatie en Ondernemen}) procured a commercial license key to conduct pilot testing and technical trials of the technology. \\
\textbf{Institutional Demonstrations:} Field evaluations and public-sector de-identification workflows were prototyped and demonstrated by Rik Doclo within the Flemish Department of Environment and Spatial Planning (\emph{Departement Omgeving, Flanders}). \\
\textbf{Open-Source Availability \& Permanent Archival:} The native Java SDK is publicly available on GitHub at \url{https://github.com/rikdoclo/wowool-java-sdk}. The official academic publication and archival record is permanently indexed on Zenodo under DOI: \href{https://doi.org/10.5281/zenodo.23441610}{\texttt{10.5281/zenodo.23441610}} with European OpenAIRE integration. \\
\textbf{Competing Interests:} Philippe Forest, Djamel Grine, and Carolina Rubio De Hita are employees or shareholders of Wowool AB. Rik Doclo is an independent researcher acting as an external contributor to the Java SDK ecosystem.

\begingroup
\small
\setlength{\itemsep}{1pt plus 0.5pt minus 0.5pt}
\bibliographystyle{plain}
\bibliography{references}
\endgroup

\end{document}